\documentclass[conference]{IEEEtran}
\IEEEoverridecommandlockouts
\usepackage{cite}
\usepackage{amsmath,amssymb,amsfonts}
\usepackage{algorithmic}
\usepackage{graphicx}
\usepackage{textcomp}
\usepackage{xcolor}
\usepackage{hyperref}
\def\BibTeX{{\rm B\kern-.05em{\sc i\kern-.025em b}\kern-.08em
    T\kern-.1667em\lower.7ex\hbox{E}\kern-.125emX}}
\begin{document}

\title{An Explorative Analysis of SVM Classifier and ResNet50 Architecture on African Food Classification\\

}

\author{
\IEEEauthorblockN{
\textsuperscript{1}Chinedu Mbonu, 
\textsuperscript{2}Kenechukwu Anigbogu, 
\textsuperscript{3}Doris Asogwa,
\textsuperscript{4}Tochukwu Belonwu
}
\IEEEauthorblockA{
\textsuperscript{1}Department of Computer Science, Nazarbayev University, Astana\\
\textsuperscript{1,2,3,4}Department of Computer Science, Nnamdi Azikiwe University, Awka\\
\{ce.mbonu [@nu.edu.kz,@unizik.edu.ng], ksy.anigbogu@unizik.edu.ng, dc.asogwa@unizik.edu.ng, ts.belonwu@unizik.edu.ng\}
}

}
\IEEEaftertitletext{\vspace{-0.5cm}}  % Adjust as needed

\maketitle

\begin{abstract}
Food recognition systems has advanced significantly for Western cuisines, yet its application to African foods remains underexplored. This study addresses this gap by evaluating both deep learning and traditional machine learning methods for African food classification. We compared the performance of a fine-tuned ResNet50 model with a Support Vector Machine (SVM) classifier. The dataset comprises 1,658 images across six selected food categories that are known in Africa. To assess model effectiveness, we utilize five key evaluation metrics: Confusion matrix, F1-score, accuracy, recall and precision. Our findings offer valuable insights into the strengths and limitations of both approaches, contributing to the advancement of food recognition for African cuisines.
\end{abstract}


\section{Introduction}

The importance of food to human health and well-being cannot be overstated. It plays a crucial role in sustaining life by providing the body with essential nutrients and energy needed for proper growth, development, and overall functioning ~\cite{b1}. Food image recognition is one of the most promising applications in the field of computer vision. A system capable of accurately identifying a wide variety of food items can play a key role in helping individuals monitor their eating habits and maintain a balanced diet~\cite{b2}. Image classification can be done using both traditional techniques and deep learning methods. Traditional techniques focus on extracting basic features from images, such as color, shape, and texture. These features are then used with classical machine learning algorithms like Support Vector Machines (SVM), Random Forests, or even simple Artificial Neural Networks (ANNs) to classify the images~\cite{b3}. Traditional machine learning algorithms is the standard method for food category recognition before the 2010s and the rise of deep learning and large image datasets ~\cite{b4}.

Deep Learning is a type of machine learning that uses
many hierarchical abstractions of data to provide users a
better understanding of the data ~\cite{b5}. Most deep learning approaches rely on neural network architectures, which is why these models are often called deep neural networks. Among the most widely used types of deep neural networks are convolutional neural networks (CNNs or ConvNets). CNNs are especially effective for handling two-dimensional data like images because they use 2D convolutional layers to process the input. Instead of requiring manual feature extraction, CNNs automatically learn and extract useful features from the images themselves, making them a powerful tool for image classification and other computer vision tasks ~\cite{b6, b7}. Image classification has made remarkable progress with the help of pre-trained CNN models like VGG16, DenseNet, MobileNet, Inception V3, ResNet50, and Xception. These models have played a key role in improving the accuracy and performance of computer vision systems~\cite{b8}.

Previous studies have explored the challenging task of food image classification using both traditional machine learning (ML) and deep learning (DL) approaches. Traditional ML methods employed in this domain include the use of color histograms and Gabor filters for feature extraction, the integration of individual dietary patterns with image analysis outputs, the application of neural networks, support vector machine (SVM)-based image segmentation techniques, and the use of food images from specific domains or categories ~\cite{b9}.
Classifying food images is particularly difficult because photos of the same dish can look very different. This happens due to differences in how people take pictures, such as the angle, lighting, or even how the food is arranged. Also, everyday items such as utensils, bottles, or glasses often appear in the background, adding more complexity to the task~\cite{b10}.

Our study makes the following key contributions to the existing knowledge in this domain:

\begin{itemize}
\item It presents a comparative analysis of deep learning and classical machine learning models on African food datasets.

\item It goes beyond overall accuracy by providing a detailed, per-class performance analysis, offering a more granular view of model behavior across different food categories.

\item Unlike earlier studies that didn’t share their code, we’ve made our full codebase available to help others easily reproduce and build on our work.
\end{itemize}

\subsection{Dataset}
\begin{figure}[h] % "h" means here
    \centering
    \includegraphics[width=0.5\textwidth]{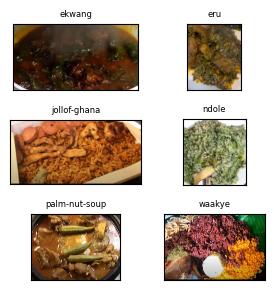} 
    \caption{Sample visualization of Each Class in our Dataset}
    \label{foodsample}
\end{figure}

The main source of our dataset is African foods datasets\footnote{\url{https://data.mendeley.com/datasets/rrzhwbg3kw/}}. It consists of six classes: Ekwang, Eru, Ndole, Jollof Rice, Palm Nut Soup, and Waakye, as illustrated in Fig.~\ref{foodsample}. It contains a total of 1,658 samples, though the distribution across classes are uneven, resulting in an imbalanced dataset. For training and evaluation purposes, the dataset was divided into three parts: 70\% for training, 15\% for testing, and 15\% for validation, as shown in Table~\ref{datsetdistribution}. 
\begin{table}[h]
    \centering
    \begin{tabular}{lccc}
        \hline
        \textbf{Class} & \textbf{Train} & \textbf{Validation} & \textbf{Test} \\
        \hline
        Ekwang & 142 & 30 & 31 \\
        Eru & 145 & 30 & 31 \\
        Jollof Rice (Ghana) & 243 & 52 & 52 \\
        Ndole & 144 & 30 & 31 \\
        Palm Nut Soup & 274 & 59 & 59 \\
        Waakye & 214 & 45 & 47 \\
        \hline
        \textbf{Total} & 1162 & 246 & 251 \\
        \hline
    \end{tabular}
    \caption{Original Dataset Split: Training, Validation, and Test Sets}
    \label{datsetdistribution}
\end{table}

The selected food are popular dishes in some part of african; Eru is a traditional dish popularly eaten in
Cameroon and neighbouring Nigeria ~\cite{b21}. Ekwang and Ndole are also well-known dishes in Cameroon, while Jollof Rice, Waakye, and Palm Nut Soup are Ghanian food. ~\cite{b22}.

\subsection{Related Works}
\subsubsection {Deep Learning Approach}
Advancements in deep learning, coupled with the availability of large-scale datasets and powerful computational resources, have simplified the process of image classification. Convolutional Neural Networks (CNNs), a class of deep learning network have emerged as the most popular and widely adopted method for image classification in recent times. In particular, applying image classification to diverse food-related datasets has seen significant improvements through the use of various transfer learning strategies, which help optimize model performance and efficiency ~\cite{b11}. Architectures like AlexNet, VGG, ResNet, and MobileNet are commonly used in food image classification ~\cite{b12}. Transfer learning, where models pre-trained on large datasets (e.g., ImageNet) are fine-tuned for specific food datasets, has become a standard practice to improve performance and reduce training time. Data augmentation techniques, such as geometric transformations and image manipulations, are also crucial to enhance the system's performance, especially when dealing with limited datasets ~\cite{b13}.

While food image classification has been widely explored, research focusing specifically on African food image classification is still emerging. ~\cite{b14} contributed to this area by developing a CNN model specifically for classifying Yoruba foods, addressing the lack of datasets familiar to the Nigerian system. ~\cite{b15} proposed using CNNs to classify and recommend healthy diets in West Africa. They highlighted the importance of addressing improper dietary habits in the region and aimed to provide a system that recommends healthy food combinations based on nutritional value and food availability.

Deep learning models, particularly Convolutional Neural Networks, have significantly advanced food image classification across various domains. Dietary assessment systems like NutriNet ~\cite{b16} utilize architectures such as Google's Inception to recognize food and drink images. For classifying Korean foods, researchers fine-tune pre-trained models on public datasets to enhance accuracy ~\cite{b9}. In Asian food classification, models like MobileNet, VGG, and ResNet are employed, with techniques such as comparing correlation coefficients to address intra-class differences ~\cite{b12}. Common techniques include transfer learning, where models pre-trained on large datasets are adapted for food datasets, and data augmentation, which enhances model generalization ~\cite{b13}.

\subsubsection {Classical Machine Learning Approach}
Classical machine learning techniques have played a foundational role in early food image recognition systems, particularly before the rise of deep learning architectures. ~\cite{b17} introduced a food classification framework that utilized multiple kernel learning-based Support Vector Machines (SVM) combined with color histograms and texture features. This approach effectively captured both chromatic and spatial patterns from food images, enabling the classification of a diverse set of Japanese dishes. Similarly, ~\cite{b18} demonstrated the utility of handcrafted features such as Histogram of Oriented Gradients (HOG) for image classification. Their study employed SVMs to distinguish visual patterns in food datasets, providing an efficient pipeline that required significantly less computational overhead compared to modern deep learning models.

Other studies also explored alternative non-deep learning strategies to achieve robust food classification. ~\cite{b19} proposed a method that detected candidate food regions using global image descriptors and classified them using the k-Nearest Neighbors (kNN) algorithm. This approach proved particularly effective for multi-dish scenarios where food items appeared in cluttered backgrounds. In a related work, ~\cite{b20} applied Random Forest classifiers on Bag-of-Features representations, leveraging the ensemble nature of the model to improve robustness across diverse food categories. These studies collectively highlight the relevance of classical methods in contexts with limited data, where deep networks may overfit or require extensive computational resources.

\section{Methodology}

\begin{figure}[h] % "h" means here
    \centering
    \includegraphics[width=0.5\textwidth]{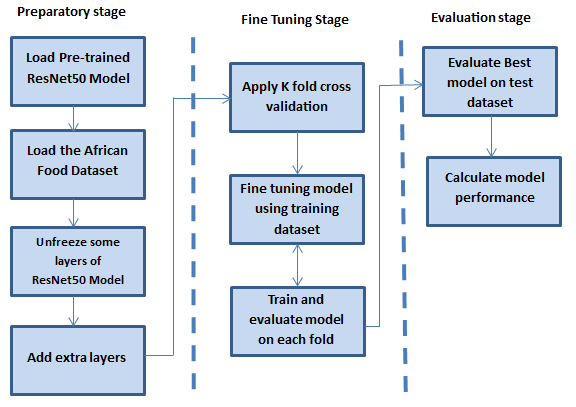} 
    \caption{Methodology for Fine-Tuning ResNet50 model on African food dataset}
    \label{method}
\end{figure}

\subsection{ResNet50 Architecture}
\begin{figure}[h] % "h" means here
    \centering
    \includegraphics[width=0.5\textwidth, height = 4cm]{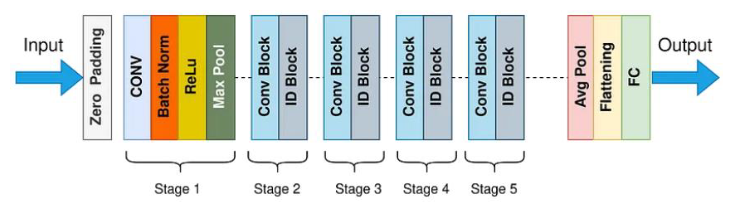} 
    \caption{ResNet50 Architecture ~\cite{b28}}
    \label{resneta}
\end{figure}

ResNet-50, with 50 layers is one of the variants of Residual Network(ResNet), a convolutional neural network (CNN). It has 48 Convolution
layers along with 1 MaxPool and 1 Average Pool layer as shown in Fig.~\ref{resneta}. Despite having 50 layers, it has more than 23 million trainable parameters ~\cite{b23}. It is a prebuilt model which has been trained on the ImageNet dataset for identifying different images of 1,000 classes~\cite{b24}. It uses residual learning with skip (identity) connections, which allow gradients to pass straight through the layers. This makes it easier to train very deep networks effectively ~\cite{b25} In this study, we leverage transfer learning to utilize the pretrained weights of the ResNet-50 architecture. Transfer learning is often used in deep learning to apply a pretrained network to new classification or prediction tasks. To adapt the model for a different purpose, its parameters are fine-tuned by adjusting the existing weights and integrating new ones that are randomly initialized to suit the specific task. ~\cite{b26}. Our approach to fine-tuning the ResNet50 model for food classification followed three key stages: preparatory, fine-tuning, and evaluation as shown in Fig.~\ref{method}.

At the preparatory stage, we loaded our dataset along with the pre-trained ResNet50 model and discarded the fully connected layer (include\_top = false). Initially, the dataset was divided into training, validation, and testing sets by its original source. However, to align with our training strategy, we reorganized it into just training and testing sets. This adjustment was necessary because we utilized K-fold cross-validation during training. To implement this, we combined the original training and validation sets into a single training dataset. We also made modifications to the model by unfreezing the last four trainable layers of ResNet50 to enable fine-tuning. In addition, we introduced three new layers, which included:
\begin{itemize}
\item Global Average Pooling 2D Layer – Converts feature maps into a single vector to reduce complexity.
\item Fully Connected Layer – Includes 1,024 neurons with a ReLU activation function and L2 regularization (0.01) to control overfitting. A dropout rate of 0.5 was also applied to randomly deactivate neurons during training, improving generalization.
\item Final Dense Layer – Determines the classification output, with the number of neurons matching the number of food categories (6) and using softmax activation to generate probability scores.
\end{itemize}
Since ResNet50 was originally trained on 224×224 images, we resized our dataset to 224×224. Also, we applied data augmentation to expand and diversify the dataset. Since the dataset size was relatively small, augmentation helped reduce the risk of overfitting and improved the model’s ability to generalize. Data augmentation helps grow a dataset by creating altered versions of the original data through various modifications~\cite{b27}. The following augmentation techniques were applied:
\begin{itemize}
\item Normalize pixel values (1./255)
\item Randomly rotate images vertically (0.2)
\item Shear transformation (0.2)
\item height\_shift\_range(0.2)
\item Random zoom (0.2)
\item Flip images horizontally (True)
\end{itemize}

During the fine-tuning stage, we used 5-fold cross-validation, where the model was trained separately on each fold. We manually selected and applied different hyperparameter tuning in a bid to get better results. After training, we saved all the models and selected the one with the highest validation accuracy as the best-performing model. We selected these hyperparameters for the final model training because they provided better accuracy:
\begin{itemize}
\item Batch size (32)
\item Optimizer (Adam)
\item Learning rate (0.0001)
\item Number of epochs (100)
\end{itemize}
During the evaluation stage, we used the test dataset to evaluate the model that performed best during K-fold cross-validation. The evaluation metrics used are listed in Table~\ref{tab:metrics_equations}.
The code for fine-tuning ResNet50 Architecture on African Foods Datasets is available on GitHub\footnote{\url{https://github.com/NedumCares/AfricanFoodClassificationResnet50}}.

\subsection{Support Vector Machine Classifier}
\begin{figure}[h] % "h" means here
    \centering
    \includegraphics[width=0.5\textwidth, height =7cm]{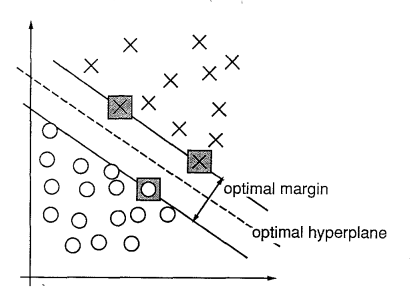} 
    \caption{SVM Classification with Support Vectors and Optimal Hyperplane} ~\cite{b30}
    \label{SVMCSVOH}
\end{figure}

Support Vector Machines are used for classification and regression; it belong to generalized linear classifiers.  SVM is a method mostly used in pattern and object recognition ~\cite{b29}. The objective of the support vector machine as shown in Fig ~\ref{SVMCSVOH} is to form a hyperplane as the decision surface so that the margin of separation between positive and negative examples is maximized by utilizing optimization approach. Generally, linear functions are used as a separating hyperplane in the feature space.

\begin{figure}[h] % "h" means here
    \centering
    \includegraphics[width=0.5\textwidth, height =6cm]{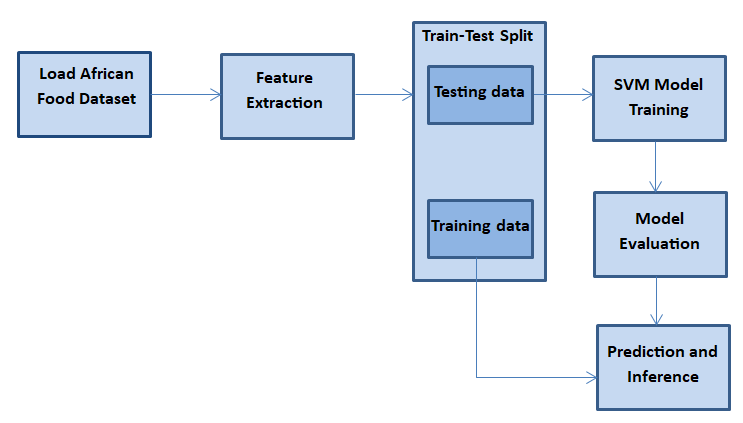} 
    \caption{Methodology overview for African food classification using SVM classifier} 
    \label{SVM method}
\end{figure}

At the initial stage of African food classification with the SVM classifier, we loaded the African food image dataset and performed feature extraction using the raw pixel intensities approach. We used the RGB values of each image as input features for the Support Vector Machine (SVM) classifier. All images were resized to a fixed size of 100×100 pixels, while retaining their original three color channels. After resizing, each image was flattened into a one-dimensional feature vector of 30,000 elements (100 × 100 × 3), representing the raw RGB pixel values. This approach offers a straightforward and computationally efficient way to prepare image data for classification, as it does not involve any additional preprocessing like grayscale conversion or handcrafted feature extraction. While this method allows the model to utilize both color and spatial information, it may also be more sensitive to variations in lighting, background, and image quality.

At the train-test split stage, we combined the original training and validation sets provided with the dataset into a single training set. The test dataset supplied by the source our dataset was kept unchanged and used exclusively for the final evaluation of the model.

For the model training phase, we used a Support Vector Machine (SVM) classifier to predict the six classes in the African foods dataset. The model was trained using the hyperparameter settings provided by the scikit-learn SVC implementation. The following values were used for the the hyperparameters during the model training:

\begin{itemize}
\item Kernel: Radial Basis Function(RBF)
\item C: 1.0
\item Gamma: Scale
\item Degree: 3
\end{itemize}

The model was evaluated using the specified evaluation metrics presented in Table~\ref{tab:metrics_equations}, and its performance was assessed on the testing dataset.

\subsection{Evaluation metrics}

Evaluation metric plays a critical role in achieving the optimal classifier during the classification training. Thus, a selection of suitable evaluation metric is an important key for discriminating and obtaining the optimal classifier ~\cite{b31}. In this study we leveraged five evaluation metrics: Accuracy, Precision, Recall, F1 Score and Confusion Matrix. The metrics and it's equations are shown in Table~\ref{tab:metrics_equations}. The equation includes four important terms: True Positive (TP), False Positive (FP), True Negative (TN), and False Negative (FN). True Positives occur when the model correctly identifies a positive case, while False Positives happen when the model incorrectly labels a negative case as positive. True Negatives refer to correctly identified negative cases, whereas False Negatives occur when the model mistakenly classifies a positive case as negative.~\cite{b32}.
\begin{table}[h]
    \centering
    \renewcommand{\arraystretch}{1.3} % Adjust row height
    \begin{tabular}{|l|l|}
        \hline
        \textbf{Metrics} & \textbf{Equation} \\
        \hline
        Accuracy & \( \frac{TP + TN}{TP + TN + FP + FN} \) \\
        \hline
        Precision & \( \frac{TP}{TP + FP} \) \\
        \hline
        Recall (sensitivity) & \( \frac{TP}{TP + FN} \) \\
        \hline
        F1 Score & \( \frac{2 \times (\text{Precision} \times \text{Recall})}{\text{Precision} + \text{Recall}} \) \\
        \hline
        Confusion Matrix &       \\ 
        \hline
    \end{tabular}
    \caption{Evaluation Metrics For African food classification}
    \label{tab:metrics_equations}
\end{table}
The confusion matrix is a crosstable that records the number of occurrences between two raters, the true/actual
classification and the predicted classification. The classes are organized in the same sequence for both rows and columns, so the correctly classified cases align along the main diagonal from the top left to the bottom right. These diagonal values indicate how often the two raters made the same judgment~\cite{b33}.

\section{Experimental Results and Discussion}

\subsection{Fine-tuned ResNet50 Model Results}

\begin{figure}[h] % "h" means here
    \centering
    \includegraphics[width=0.5\textwidth, height =8cm]{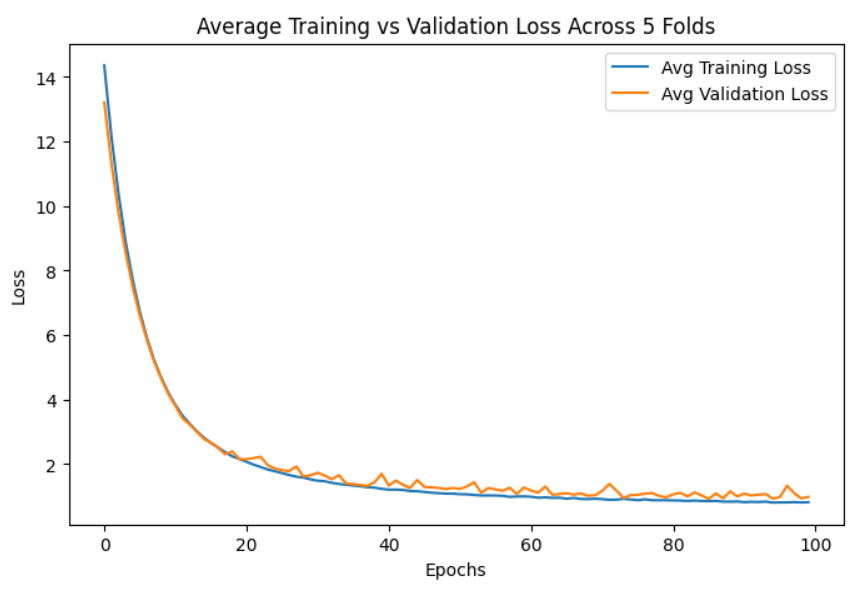} 
    \caption{Average Training vs Validation Loss Across 5 Folds in Fine-tuned ResNet50 Architecture}
    \label{losscurve}
\end{figure}

The fine-tuned ResNet50 model was evaluated using five key metrics: Confusion matrix, F1-score, accuracy, recall and precision. To confirm that the model neither overfits nor underfits, we examined the average training and validation loss curve shown in Fig.~\ref{losscurve}. The steady decline in loss over the epochs suggests that the model is effectively learning from the data.

\begin{figure}[h] % "h" means here
    \centering
    \includegraphics[width=0.5\textwidth]{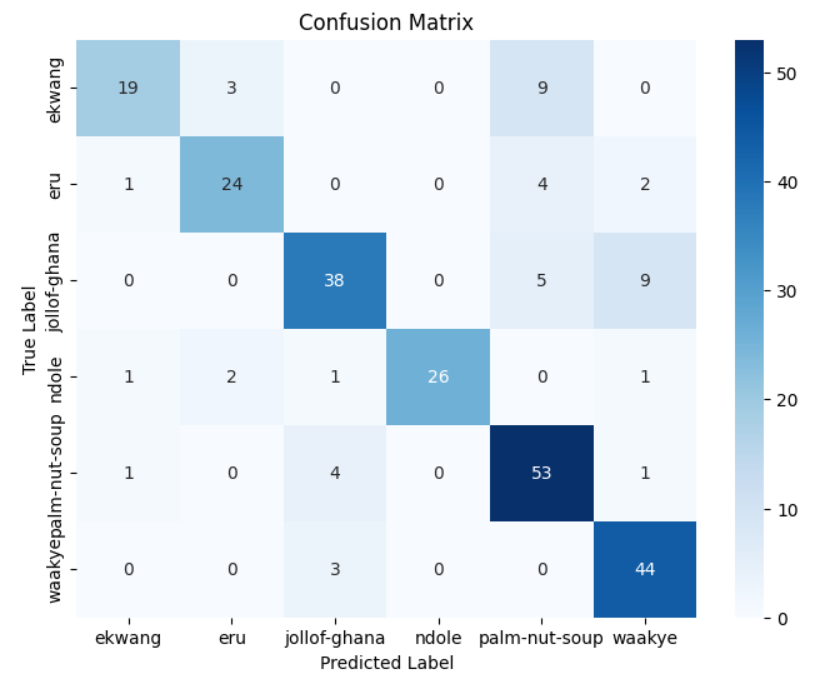} 
    \caption{Confusion Matrix evaluation report on fine-tuned ResNet50 Architecture}
    \label{cmreport}
\end{figure}

The Confusion matrix evaluation report in fig.~\ref{cmreport} illustrates how well the fine-tuned ResNet50 model performed on the test dataset. It shows both the accurate predictions and the areas where the model made errors. One notable misclassification was Palm-nut soup being incorrectly identified as Ekwang 9 times out of 31 occurrences. However, the model demonstrated strong performance in recognizing Ndole correctly by not misclassifying any other food as Ndole.

Based on the data presented in Table`\ref{tab:svm_resnet} our model achieved an accuracy of 81\% across 251 samples, indicating strong overall performance. It performed exceptionally well in identifying Ndole, with a perfect precision score of 1.00 and an F1-score of 0.91, suggesting very few false positives. However, the model struggled with Ekwang, which had the lowest recall (0.61) and F1-score (0.72), meaning it had difficulty correctly identifying all true instances. On the other hand, Waakye had the highest recall (0.94), showing that nearly all actual instances were correctly classified. Jollof-Ghana and Eru demonstrated moderate performance based on their F1-scores, indicating a balance between precision and recall.

\subsection{Support Vector Machine Classifier Results}

\begin{figure}[h] % "h" means here
    \centering
    \includegraphics[width=0.5\textwidth, height =8cm]{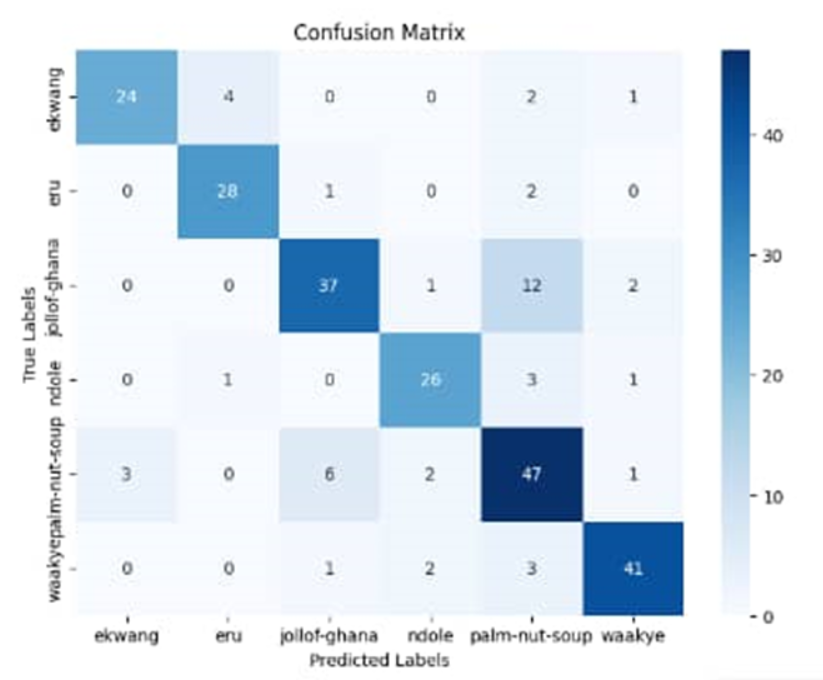} 
    \caption{Confusion Matrix evaluation report on SVM classifier}
    \label{matrics_equation}
\end{figure}

The Confusion matrix evaluation report in Fig. ~\ref{matrics_equation} illustrates how well the SVM model performed on the test dataset. The overall accuracy of the model is 80. 88\%, showing that the model performs well across the board, but there is still room to improve it, especially by focusing on the weaker classes.

The SVM classification report in Table ~\ref{tab:svm_resnet} is showing that the model's precision is generally high. All classes except palm-nut soup (0.68) are above 82\% meaning the predictions made for those classes are mostly correct. Palm-nut soup likely has some false positives. The recall shows how well the model can identify all the actual instances of each class. Most of the food labels have a recall above 80\%, showing the model is doing well in finding the correct items. Only ekwang (0.77) and jollof-ghana (0.71) are slightly below, meaning the model misses some true cases.

\subsection{Comparative Analysis of SVM and ResNet50 Architecture}

\begin{figure}
    \centering
    \includegraphics[width=0.5\textwidth, height =8cm]{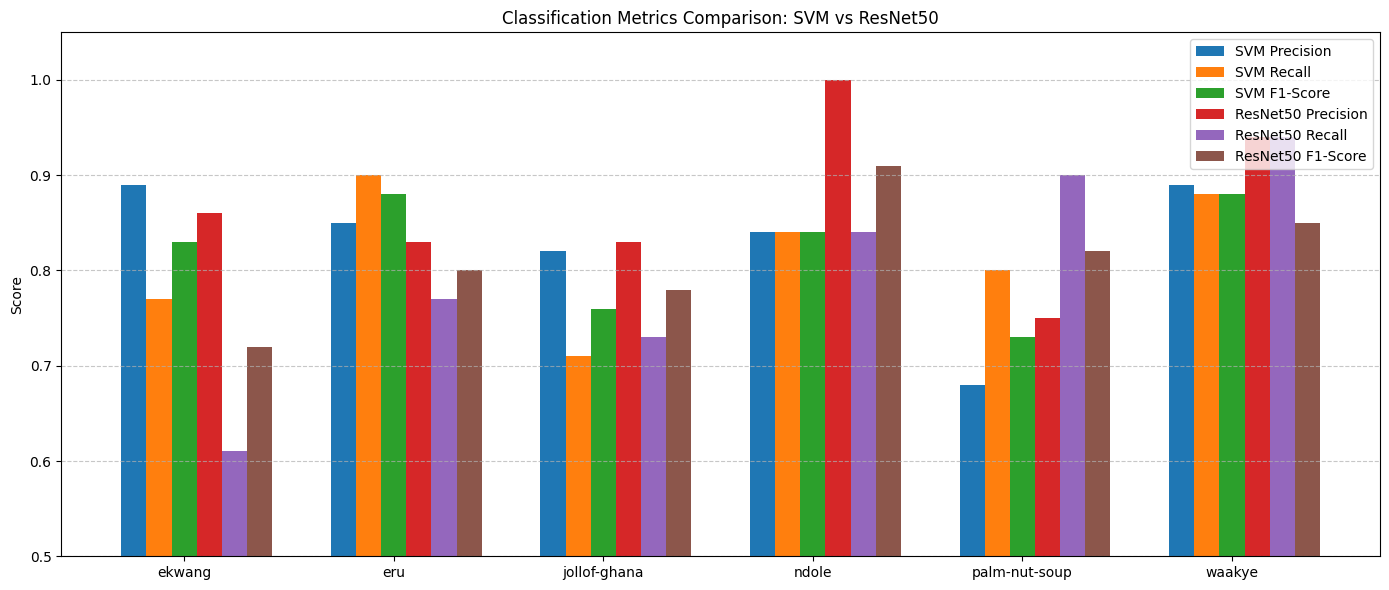} 
    \caption{Comparative analysis of performance metrics for ResNet50 and SVM}
    \label{COMPAANA}
\end{figure}

Table~\ref{tab:svm_resnet} and Fig~\ref{COMPAANA} present the performance comparison between ResNet50 Architecture and the SVM classifier on the African Foods Dataset. Both models achieved an overall accuracy of 81\%, but their effectiveness varied across different food categories. ResNet50 recorded a macro-averaged and weighted F1-score of 0.81, while SVM performed better with scores of 0.82 and 0.81, respectively. This suggests that SVM maintained more consistency across all food classes compared to ResNet50.

Examining individual classes, SVM demonstrated stronger performance in Ekwang and Eru, outperforming ResNet50 in Precision, Recall, and F1-score. ResNet50 showed a clear advantage in identifying Palm-nut-soup, achieving a Recall of 0.90, and SVM reached only 0.80. Despite these variations, both models achieved the same accuracy rate.

\begin{table}[h]
    \centering
    \scriptsize
    \resizebox{0.5\textwidth}{!}{%
    \begin{tabular}{lcccccc}
        \hline
        \textbf{Class} & \multicolumn{3}{c}{\textbf{SVM}} & \multicolumn{3}{c}{\textbf{ResNet50}} \\
        \cline{2-4} \cline{5-7}
        \textbf{Name} & Precision & Recall & F1-score & Precision & Recall & F1-score \\
        \hline
        ekwang & 0.89 & 0.77 & 0.83 & 0.86 & 0.61 & 0.72 \\
        eru & 0.85 & 0.90 & 0.88 & 0.83 & 0.77 & 0.80 \\
        jollof-ghana & 0.82 & 0.71 & 0.76 & 0.83 & 0.73 & 0.78 \\
        ndole & 0.84 & 0.84 & 0.84 & 1.00 & 0.84 & 0.91 \\
        palm-nut-soup & 0.68 & 0.80 & 0.73 & 0.75 & 0.90 & 0.82 \\
        waakye & 0.89 & 0.87 & 0.88 & 0.77 & 0.94 & 0.85 \\
        \hline
        Accuracy & \multicolumn{3}{c}{81\%} & \multicolumn{3}{c}{81\%} \\
        Macro Avg & 0.83 & 0.82 & 0.82 & 0.84 & 0.80 & 0.81 \\
        Weighted Avg & 0.82 & 0.81 & 0.81 & 0.82 & 0.81 & 0.81 \\
        \hline
    \end{tabular}%
    }
    \caption{Comparison of classification performance between SVM and ResNet50 for African food recognition.}
    \label{tab:svm_resnet}
\end{table}

\section{Conclusion}
In this study, we compared two different methods for classifying African food images: a Support Vector Machine (SVM) and a fine-tuned ResNet50 model. Although both approaches achieved the same overall accuracy of 81\%, their strengths varied across specific food categories. The ResNet50 model, which took advantage of transfer learning and data augmentation, showed strong performance in recognizing dishes such as Palm Nut Soup and Ndole, likely due to its ability to capture complex features through deep layers. On the other hand, the SVM model delivered more consistent results across the board, outperforming ResNet50 on items such as Ekwang and Jollof-Ghana. These results suggest that while deep learning models like ResNet50 offer advantages in extracting detailed visual patterns, especially when supported by data augmentation, traditional machine learning methods like SVM can still hold their ground. SVM's stability may offer better generalization in smaller or imbalanced datasets scenarios. 
Our work adds to the limited but growing body of research focused on African food classification. By comparing two fundamentally different modeling approaches and making our code publicly available, we aim to encourage further exploration in this space. Future research could explore the use of larger, more balanced datasets, test ensemble models that blend the strengths of both SVM and deep learning, or broaden the scope to include a wider variety of African cuisines, especially those specific to particular regions.

\end{document}